\documentclass[sigconf]{acmart}
\usepackage{graphicx}
\usepackage{subcaption}
\usepackage{multirow}
\usepackage{float}
\usepackage{comment}

\AtBeginDocument{%
  }

\begin{document}

\title{Fingerprinting Deep Learning Models via Network Traffic Patterns in Federated Learning}

\author{Md Nahid Hasan Shuvo}
\email{mshuvo@gmu.edu}
\orcid{0000-0003-0144-1601}
\affiliation{%
  \institution{George Mason University}
  \city{Fairfax}
  \state{Virginia}
  \country{USA}
}

\author{Moinul Hossain}
\email{mhossa5@gmu.edu}
\orcid{0000-0002-1121-7649}
\affiliation{%
  \institution{George Mason University}
  \city{Fairfax}
  \state{Virginia}
  \country{USA}
}

\thanks{This is the author's version of the work. It has been accepted for publication in the Proceedings of the 2025 ACM Workshop on Wireless Security and Machine Learning (WiseML 2025), July 3, 2025, Arlington, VA, USA. The final version will be available in the ACM Digital Library.}

%\renewcommand{\shortauthors}{Md Nahid Hasan Shuvo and Moinul Hossain}

%%
%% The abstract is a short summary of the work to be presented in the
%% article.
\begin{abstract}
    Federated Learning (FL) is increasingly adopted as a decentralized machine learning paradigm due to its capability to preserve data privacy by training models without centralizing user data. However, FL is susceptible to indirect privacy breaches via network traffic analysis—an area not explored in existing research. The primary objective of this research is to study the feasibility of fingerprinting deep learning models deployed within FL environments by analyzing their network-layer traffic information. In this paper, we conduct an experimental evaluation using various deep learning architectures (i.e., CNN, RNN) within a federated learning testbed. We utilize machine learning algorithms, including Support Vector Machines (SVM), Random Forest, and Gradient-Boosting, to fingerprint unique patterns within the traffic data. Our experiments show high fingerprinting accuracy, achieving 100\% accuracy using Random Forest and around 95.7\% accuracy using SVM and Gradient Boosting classifiers. This analysis suggests that we can identify specific architectures running within the subsection of the network traffic. Hence, if an adversary knows about the underlying DL architecture, they can exploit that information and conduct targeted attacks. These findings suggest a notable security vulnerability in FL systems and the necessity of strengthening it at the network level.
\end{abstract}



%%
%% Keywords. The author(s) should pick words that accurately describe
%% the work being presented. Separate the keywords with commas.
%% A "teaser" image appears between the author and affiliation
%% information and the body of the document, and typically spans the
%% page.

%\received{20 February 2007}
%\received[revised]{12 March 2009}
%\received[accepted]{5 June 2009}

%%
%% This command processes the author and affiliation and title
%% information and builds the first part of the formatted document.
\maketitle

\section{Introduction}
Federated learning (FL) has improved the training process of artificial intelligence (AI) models by allowing multiple devices to collaborate without sharing raw data \cite{mcmahan2017communication}. This method limits data access, hence improving privacy; it is especially advantageous in fields like the Internet of Things (IoT) networks, autonomous robotics,  and healthcare \cite{kairouz2021advances}. Federated learning (FL) primarily shares model updates and enables local model training, hence obviating the necessity of transmitting private information to a centralized server. However, FL is not immune to security concerns despite its privacy-preserving characteristics. Although extensive research has concentrated on direct attacks, such as data poisoning, membership inference attacks, and model inversion, there has been little focus on indirect privacy concerns, especially vulnerabilities associated with network traffic analysis \cite{kaushal2025securing},\cite{ zhang2024survey}.
%\cite{sheller2020federated}

Federated learning showcases unique features, particularly its dependency on clients and servers for sharing model updates throughout training phases. Although these updates are encrypted, their network traffic features (e.g., packet sizes, transmission direction, and interarrival time) may reveal private information. Previous studies have demonstrated that utilizing network traffic makes it possible to do device fingerprinting in federated learning. Consequently, an attacker could actively monitor and identify the unique devices from their communication patterns \cite{he2021edge}. However, a critical and unexplored issue is whether one can fingerprint the deep learning models by analyzing network traffic patterns in FL. 

Due to the different computation properties of deep learning (DL) architectures in FL, Convolutional Neural Networks (CNN) and Recurrent Neural Networks (RNN) utilize system resources differently, especially when sending data over networks. At present, there is no prior research exploring whether FL traffic leakage information can be used to fingerprint the DL models being trained in the FL systems. This research gap is critical because if an adversary can identify these neural network structures from network leakage, then it will enable the malicious actor to develop and launch highly effective architecture-specific targeted cyberattacks, such as creating adversarial attacks explicitly designed for CNNs \cite{zhou2024stealthy} and RNNs \cite{zhang2024local}. Thereby, this study aims to address this research gap by investigating the possibility of DL architecture fingerprinting using network traffic data, such as timing information, packet size distribution, and traffic direction. If these core models in FL systems can be fingerprinted this way, it would introduce a new class of security risk, potentially undermining the privacy benefits of FL.

This study investigates the aforementioned research gap of how standard machine learning techniques can effectively fingerprint deep learning architectures deployed in federated learning environments by analyzing network layer traffic patterns. We aim to validate the feasibility of fingerprinting deep learning models using easily accessible network metadata. To achieve this goal, we perform a systematic experiment using NVIDIA GPU-equipped clients and servers running various deep-learning architectures and sniff their network traffic using a packet sniffing tool, Wireshark. We deployed popular machine-learning classification algorithms to fingerprint the DL architecture, including Random Forest, Support Vector Machine, and Gradient Boosting. Our experimental results confirm that deep learning architectures exhibit distinguishable network traffic patterns in the FL system, making them vulnerable to fingerprinting attacks. 

% Insert a paragraph outlining the contributions in this paper
In this paper, we make the following key contributions: (1) We propose a novel method to fingerprint deep learning architecture in federated learning using network-layer traffic patterns. (2) We developed a controlled FL testbed to analyze the network pattern using ideal and noisy network conditions. (3) Using traditional machine learning algorithms, we show the feasibility of this fingerprinting attack, which poses security threats to the FL environments. (4) Finally, we discuss the implications of our findings, highlight current limitations, and propose future research directions to improve both fingerprint attacks and potential defense mechanisms. \vspace{-0.16 in}

\section{Related Works}

Traffic fingerprinting (TF) is a widely studied traffic analysis approach for identifying objects such as mobile applications, website browsing, or devices according to the distinctive characteristics or behavioral patterns of network traffic. Network managers use TF for security monitoring and filtering \cite{ma2020pinpointing}, \cite{ma2021inferring}, whereas adversaries leverage TF to intercept sensitive information \cite{acar2020peek}, \cite{wang2020high}.

Website fingerprinting has garnered significant attention in the fields of web security and privacy. Despite the safeguards provided by encryption technologies, such as HTTPS and Tor, Panchenko et al. demonstrated the potential to analyze encrypted online traffic to determine the websites accessed by users \cite{panchenko2016website}. In \cite{rahman2019tik}, Rahman et al. showed that a passive local eavesdropper could utilize website fingerprinting to reveal the web browsing activities of Tor users. Furthermore, a significant area of study is device fingerprinting, which aims to recognize individual devices based on their distinct hardware and software characteristics. Laperdrix et al. examined browser-based device fingerprinting and demonstrated that seemingly benign information—such as installed plugins, screen resolution, and system fonts—can uniquely identify devices, thereby posing significant privacy concerns \cite{laperdrix2016beauty}. In \cite{sheng2025network}, researchers have extensively discussed fingerprinting industrial IoT devices from network traffic information.

On top of devices and websites, fingerprinting has been extensively applied in mobile application identification, where network flow analysis can be used to identify mobile applications. In \cite{taylor2017robust}, Taylor et al. explored how mobile apps generate unique traffic signals recognizable via network-layer data, thereby permitting attackers to determine app types and usage behaviors. In addition, Li et al. show that the application's traffic data contains multiplexed user traffic. Hence, they provide a robust app fingerprinting method \cite{li2024robust}. In \cite{yoon2024scalable}, researchers have shown that mobile activity can also be fingerprinted using control channel information in 5G networks.

While these studies focus on fingerprinting user activities and applications, a parallel research direction explores fingerprinting in distributed learning environments such as federated learning. The advancement of distributed machine learning has led to the integration of fingerprinting with FL. In \cite{melis2019exploiting}, Melis et al. showed the capability of fingerprinting an FL client device using communication patterns, which reveals the client's identification. Song et al. investigated how attackers can utilize network information in FL systems to identify and monitor specific clients \cite{song2020analyzing}. 

However, despite having tremendous progress in various domains, fingerprinting deep learning architectures in FL systems is a critical research gap. FL relies heavily on server-client communication while running different neural networks. Hence, network traffic may leak sensitive information about these models, and despite having potential security risks, no research has explicitly explored whether adversaries can fingerprint DL architectures. In this work, we explicitly ventured into this unexplored research area by analyzing network layer traffic generated by federated learning systems to fingerprint DL architectures. \vspace{-0.05 in}

\section{Methodologies}
In this section, we will discuss the experimental design, the configuration of our federated learning testbed, the procedure for collecting data, the feature engineering techniques, and the fingerprint approach that we utilize to conduct fingerprint attacks on federated learning systems. \vspace{-0.05 in}
\subsection{Experimental Design}
\begin{figure}[t]
    \centering
    \includegraphics[width=\columnwidth]{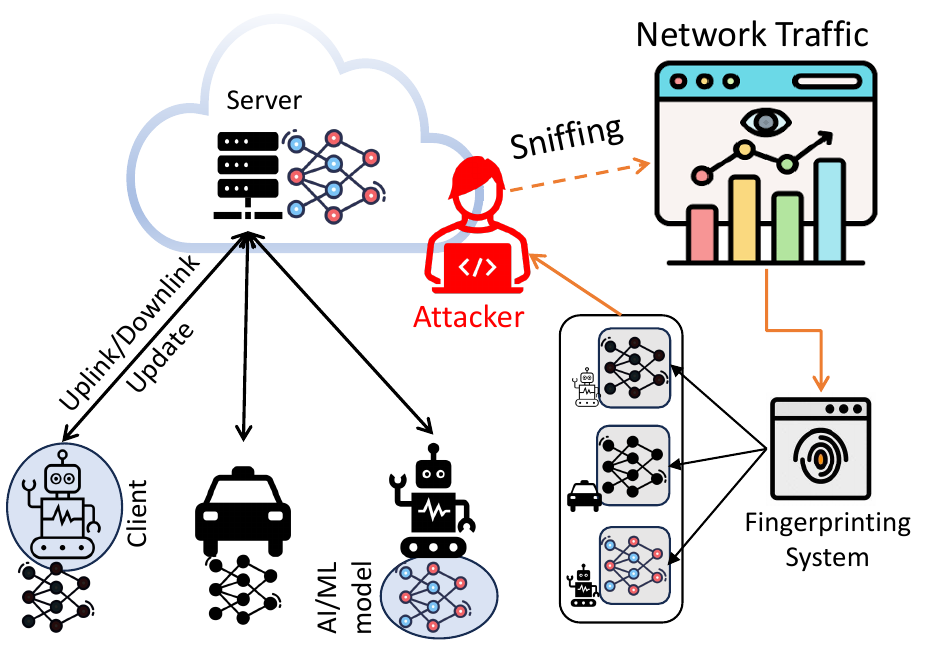}
    \caption{Framework for fingerprinting attack on FL.}
    \Description{system design }
    \label{fig:s_model}
    \vspace{-0.2cm}
\end{figure}
\vspace{-0.1cm}
The primary goal of our research is to identify various DL  architectures by evaluating network traffic data generated during federated learning. To accomplish this goal, we concentrate on the training phase of the FL framework, where multiple devices work together to train a global model in a distributed and secure manner.

A federated learning system with star topology generally contains two fundamental components: client and server. Each client individually trains its own local model, producing local updates. The updates are then sent to the central server, integrating all client updates to train a global model. The improved global model parameters are later propagated to clients, thereby triggering another cycle of local training. This repeated exchange between the server and clients continues until convergence or a specified training criterion is achieved. Federated learning depends on the network connection, making data transmission between the server and clients an extremely vulnerable point. Despite the encryption of the model parameters, the network traffic inevitably reveals important metadata, including packet sizes, transmission direction, and interarrival times. A skilled attacker with access to packet-sniffing tools can passively acquire this compromised information during the parameter update phase.

% Insert a subsection explaining the Threat Model. The threat model needs to mention that the considered adversary can extract only the Layer 3 packet size and the source and destination of the packet. It does not consider any flow-level or application-level data or control information.

Figure \ref{fig:s_model} shows a high-level overview of our system model designed for fingerprinting DL networks in FL scenarios. In this approach, the adversary intercepts network packets transmitted during FL training sessions via conventional packet-sniffing methods. The collected network data is further processed and analyzed using our proposed fingerprinting system. After processing by the fingerprinting framework, an adversary can identify both the model architecture (e.g., CNN, RNN, Transformer) and the data modality type (e.g., text, audio, video) used for training. When the attacker obtains this information, then they can launch an effective, targeted attack on those models and modalities.

\subsection{Threat Model}
In this research, we consider a passive network adversary who can monitor communication between the FL clients and the server. The adversary does not actively interfere with the FL training process but aims to fingerprint deep learning architecture based on network-layer traffic patterns. We assume the adversary has the ability to (1) sniff layer-3 traffic using widely available packet capturing tools, like Wireshark; (2) extract basic packet metadata, including packet size and packet direction; and (3) observe the communication timing patterns based on the interarrival times.
However, the adversary does not have access to the flow level or application layer data, such as payload information or FL model updates. Also, the adversary cannot access the higher-layer control information, such as the TCP sequence. Given those constraints, the adversary aims to identify distinct traffic signatures associated with the different DL architectures and infer the DL structure and its modality. This fingerprinting capability could enable further targeted adversarial attacks on the FL system by exploiting model-specific vulnerabilities.

\subsection{Testbed Setup}
\begin{figure}[t]
    \centering
    \includegraphics[width=0.9\columnwidth]{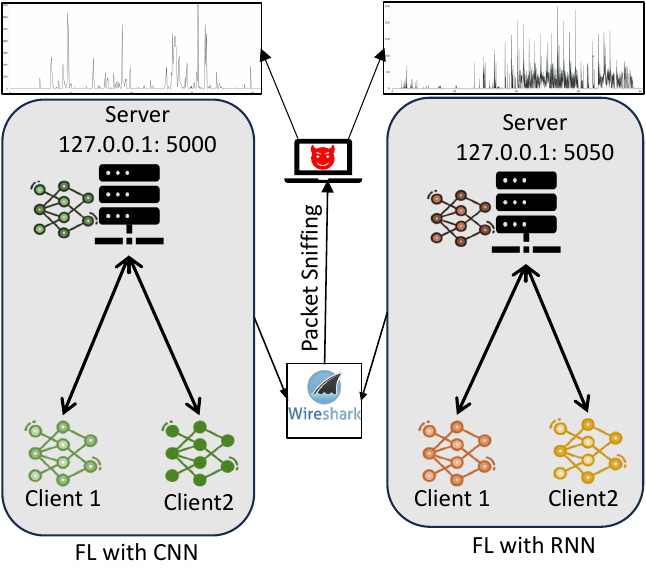}
    \caption{Testbed design for the fingerprint attack.}
    \Description{Testbed design }
    \label{fig:t_bed}
\end{figure}
In our experiment, we develop a basic federated learning testbed to assess our fingerprinting approach. To achieve this, we select two popular deep learning architectures, such as CNN and RNN, due to their fundamental differences in computational and data processing structures. CNN models are commonly used for spatial data (e.g., images and videos), while RNNs are used for sequential data with temporal features. These unique computational patterns of CNNs and RNNs introduce different training update patterns, which directly influence the communication pattern in federated learning.

Since this is a preliminary study to determine whether deep learning models are fingerprintable, we conducted our experiments using a local host setup on a single machine instead of executing training over a wireless network. The primary reason for employing a local host setup instead of a proper wireless configuration is to provide a controlled federated learning environment. We reduce outside noise by isolating external network traffic and maintaining consistency between experiments. Therefore, it helps to accurately identify variations in traffic patterns in deep learning architectures rather than network circumstances or environmental influences. However, in the future, we plan to assess the feasibility of this threat under more realistic conditions.

We conduct these experiments on a PC with an NVIDIA RTX 3060 GPU to enable faster iteration, optimize troubleshooting, and allow efficient packet capturing. Local client instances are simulated via unique ports within the localhost environment (e.g., 127.0.0.1 with unique port numbers). Then, we collected network traffic using packet sniffing tools (e.g., Wireshark). Figure \ref{fig:t_bed} illustrates the initial setup of our federated learning testbed and the data-collection methodology, focusing on the server-client layer-3 communication. The left section of this figure depicts the server aggregating updates from numerous clients executing the CNN model, while the right section exhibits clients using the RNN model. Furthermore, the upper region of the figure displays the type of packet collected by Wireshark.
\vspace{-0.3cm}
\subsection{Data Collection}

\begin{figure}[t]
    \centering
    % First Row
    \begin{subfigure}[t]{0.5\linewidth}
        \includegraphics[width=\linewidth]{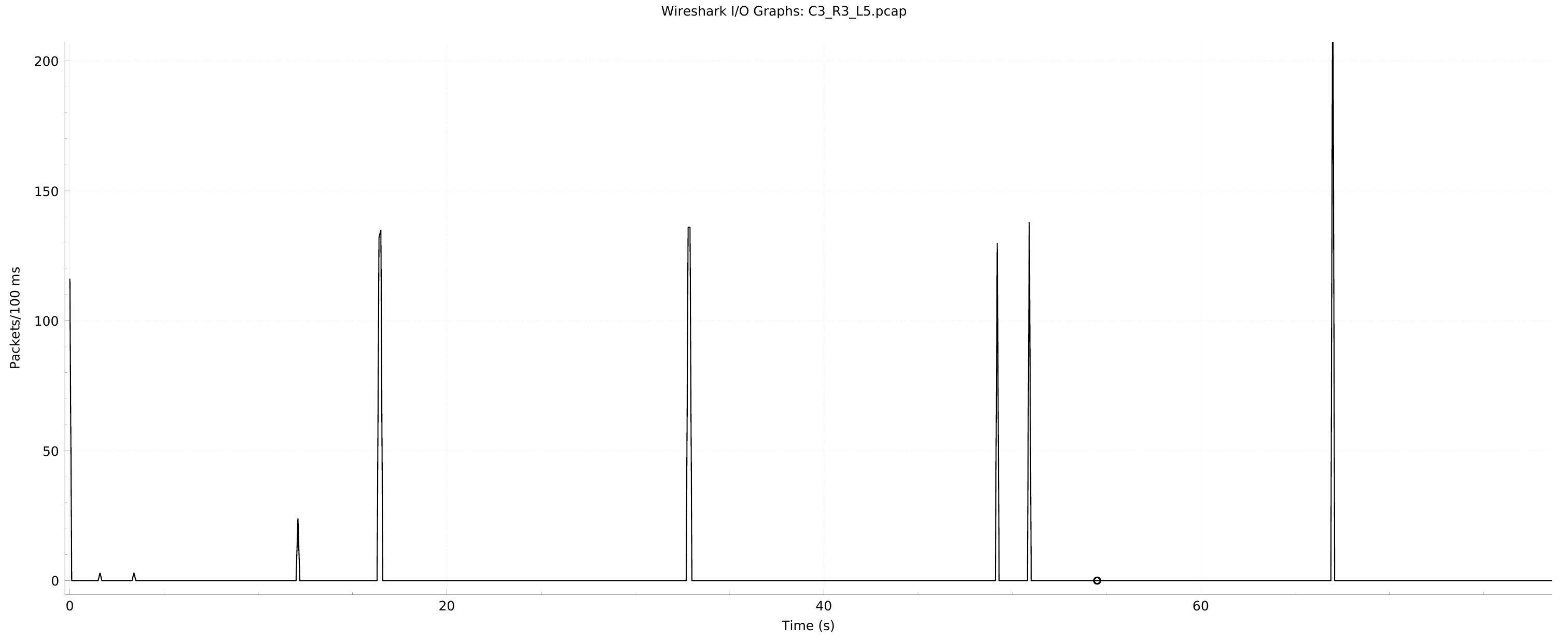}
        \caption{CNN traffic (Ideal case).}
        \label{fig:cnn_ideal}
    \end{subfigure}\hfill
    \begin{subfigure}[t]{0.5\linewidth}
        \includegraphics[width=\linewidth]{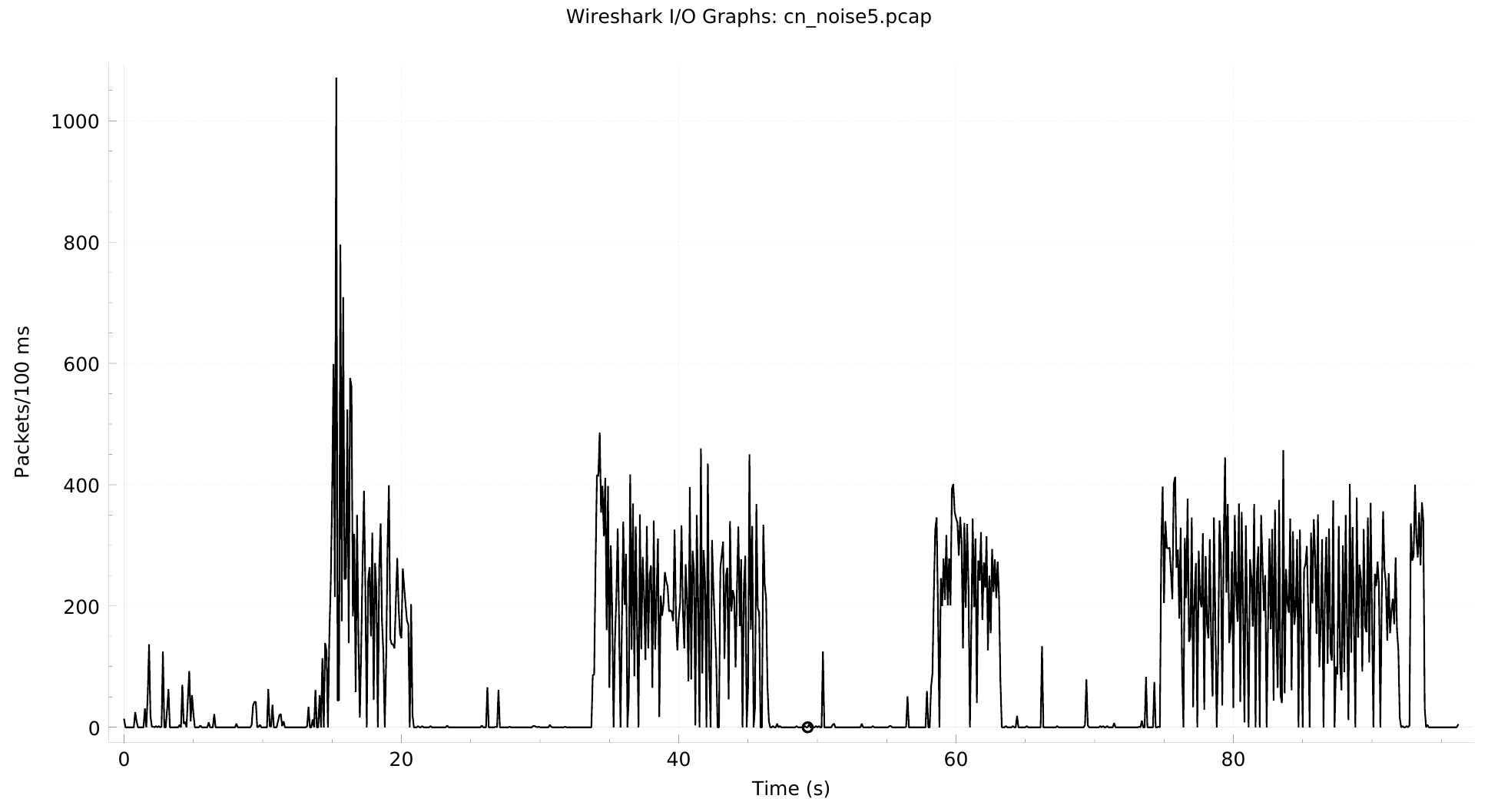}
        \caption{CNN traffic (with browsing).}
        \label{fig:cnn_noise}
    \end{subfigure}

    \vspace{0.6cm}

    % Second Row
    \begin{subfigure}[t]{0.5\linewidth}
        \includegraphics[width=\linewidth]{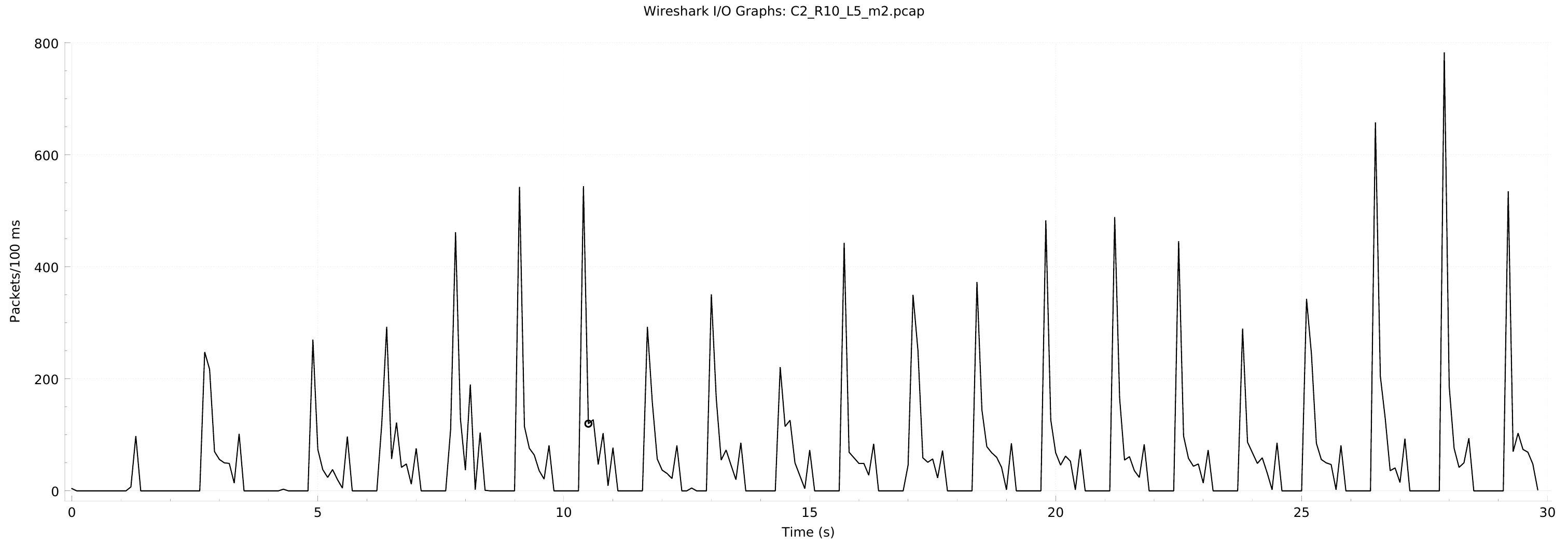}
        \caption{RNN traffic (Ideal case).}
        \label{fig:rnn_ideal}
    \end{subfigure}\hfill
    \begin{subfigure}[t]{0.5\linewidth}
        \includegraphics[width=\linewidth]{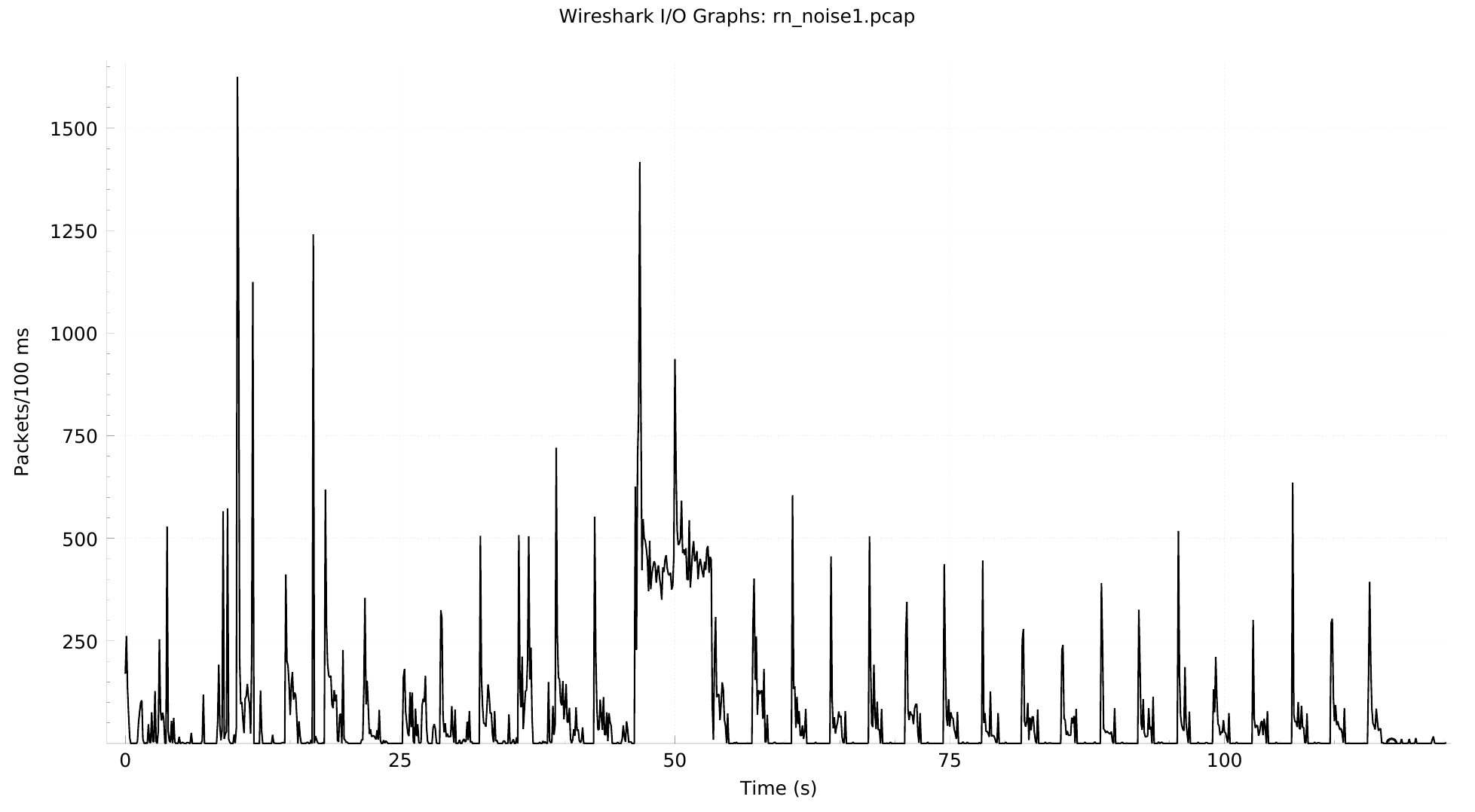}
        \caption{RNN traffic (with browsing).}
        \label{fig:rnn_noise}
    \end{subfigure}
    
    \caption{Network traffic patterns for CNN and RNN architectures under ideal and noisy experimental conditions.}
    \label{fig:traffic_2x2}
    \vspace{-0.1cm}
\end{figure}

To extensively assess the success of our fingerprinting method, we carefully collected network traffic data in controlled and realistic conditions. Data collection is conducted in two independent environments: an ideal setup with no external traffic and an alternative setting that includes external noise from internet browsing activities on clients. For each DL architecture, we use distinct datasets and tasks to ensure variational workloads.

We choose various custom CNN architectures by varying the number of layers and connections and train them on CIFAR-10 and Fashion MNIST image datasets. These databases vary in complexity, resolution, and image type. We also utilize the Sunspot Forecasting time series dataset to train customized RNN models. The reason behind choosing these customized models is to identify the patterns that relate specifically to CNN or RNN.

In order to capture variations in the dataset, we systematically modified the federated learning hyperparameters, including local and global epoch numbers and training parameters. By modifying these parameters, we change the computation of the models, hence the variation in communication patterns, which aids in obtaining fingerprint robustness and generalization.

After that, we collect the data using Wireshark, where each traffic session is saved as an individual packet capture (pcap) file. We use these PCAP files for feature extraction and analysis and then utilize those features to fingerprint the DL models. For training, we use 8 CNN and 8 RNN-based traffic data (including both ideal and noisy data) and a total of 23 PCAP files (11 for RNN, 12 for CNN) for testing our approach.

In Figure \ref{fig:traffic_2x2}, we show some collected data from both ideal and noisy cases to demonstrate how web browsing impacts network traffic.
\vspace{-0.2cm}
\subsection{Feature Engineering}
In this section, we discuss the data processing and feature selection procedures of this work.
\subsubsection{PCAP to CSV conversion}
We use Wireshark to capture raw traffic data in a PCAP file. Then we use Tshark and Python to transform the PCAP into a structured CSV file. The CSV files contained important information, including timestamps, packet length, packet direction (uplink/downlink), and interarrival times.
\begin{figure}[t]
    \centering
    \includegraphics[width=\columnwidth]{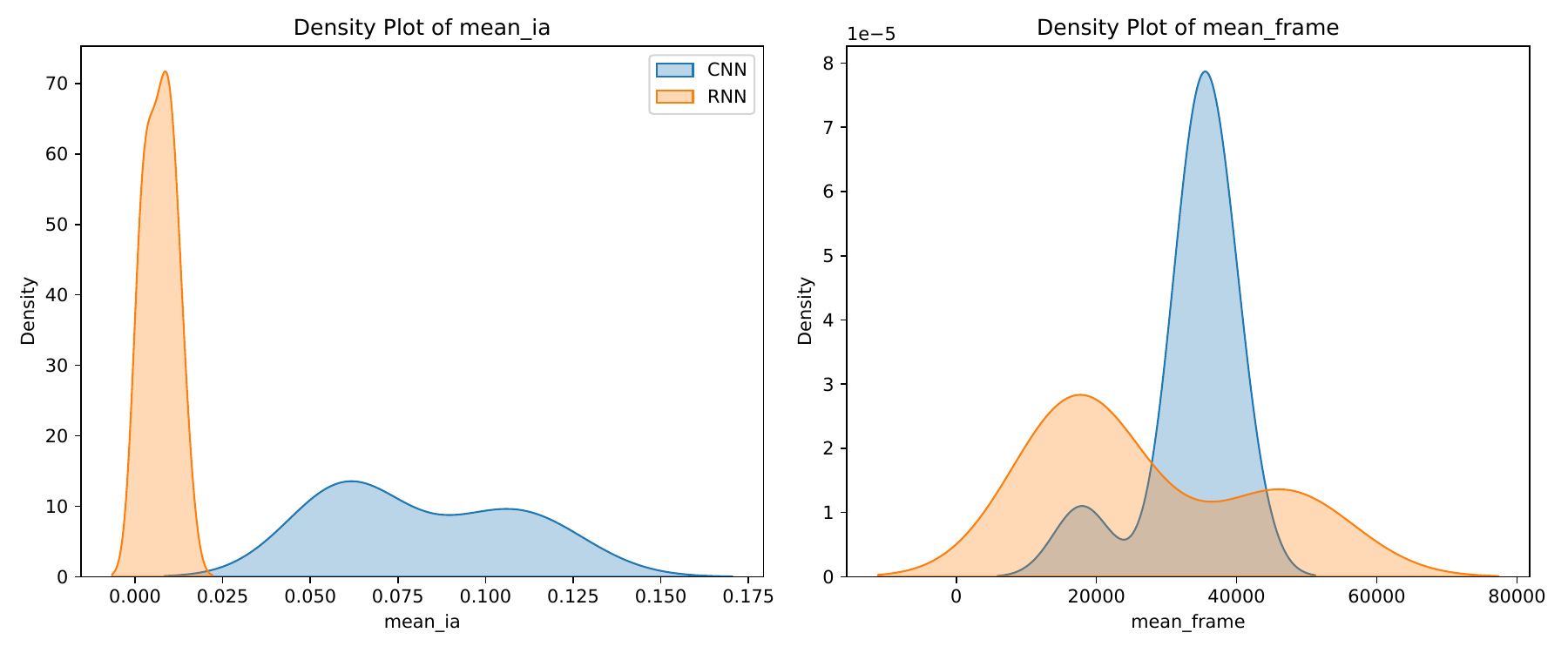}
    \vspace{-0.2cm}
    \caption{KL divergence plots of selected network traffic features (left: interarrival time; right: packet size) for CNN and RNN architectures.}
    \label{fig:kd_score}
    \vspace{-0.2cm}
\end{figure}
\vspace{-0.2cm}
\subsubsection{Derived Features}
To capture higher-level traffic behaviors, we computed statistical features such as packet-length statistics (mean, standard deviation, minimum, maximum, number of peaks), direction statistics (average direction, uplink/downlink proportions), interarrival-time statistics (mean, standard deviation, minimum, maximum, number of peaks), and overall traffic duration. However, we later discard the traffic duration because traffic duration depends on how long we are capturing the packets; it doesn't have a direct impact on the FL traffic.
We also measure the Kullback–Leibler (KL) divergence to compare CNN and RNN feature distributions. Figure \ref{fig:kd_score} shows two distributions: mean interarrival time (mean\_ia) on the left and mean packet size (mean\_frame) on the right. Strong discriminative strength is indicated by a significant KL divergence, as seen in mean\_ia, where the orange and blue curves hardly overlap. The left plot illustrates that CNN packets have more interarrival time variability, whereas RNN packets have a more concentrated distribution, indicating a more regular transmission pattern. Features having higher overlap, such as mean\_frame, have lower KL values, indicating less successful class separation. The right plot demonstrates that CNN-generated packets are larger and more uniform than RNN-generated packets, which are more dispersed. These differences show that deep learning architectures have different network traffic patterns, facilitating network-layer fingerprinting.
\vspace{-0.34cm}
\begin{figure}[H]
    \centering
    \includegraphics[width=\columnwidth]{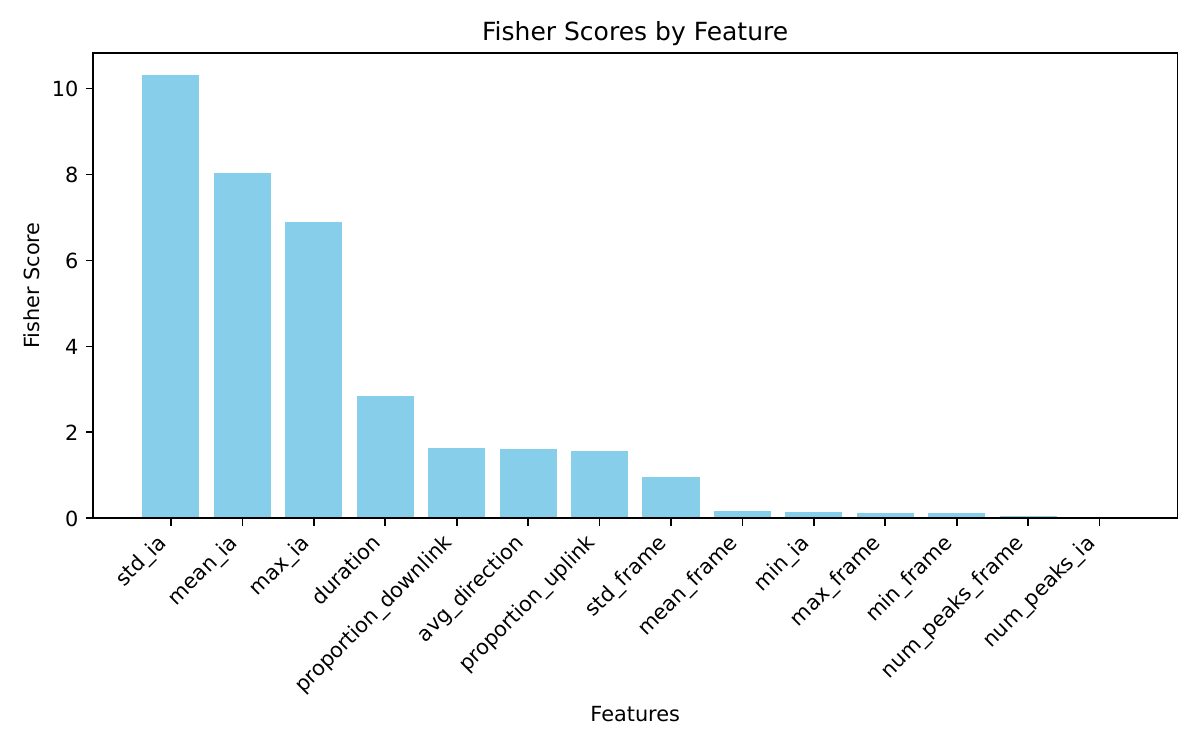}
    \vspace{-0.3cm}
    \caption{Fisher score analysis from the data features.}
    \Description{Fisher score analysis from the data features}
    \label{fig:f_score}
    %\vspace{-0.1cm}
\end{figure}
\vspace{-0.4cm}
\subsubsection{Feature Section}
Fisher score-based feature selection was employed to identify the most discriminative features. We use Fisher scores to determine the most important and discriminative features from the derived features. Fisher scores analyze the discriminative capability of features by evaluating the differences in means between classes relative to the variances within classes. Here, the classes are two: class A, which is the data from CNN training, and class B, which is from the RNN training. Figure \ref{fig:f_score} shows the computed Fisher scores for all candidate features. From here, I took the highest-ranked features for the next step. This selection mainly helps to reduce dimensionality, minimize overfitting, and increase overall classification performance.

\section{Evaluation}
In this section, we discuss the fingerprinting methods of DL models and the performance evaluation of our experiments.
\subsection{Fingerprinting Approach}
After processing the data, we selected the essential features using the methods discussed in the feature engineering section to perform the model fingerprinting. We employed well-established machine learning classifiers.

\begin{itemize}
    \item \textbf{Random Forest (RF)}:  This is an ensemble method that contains multiple decision trees trained using bagging and random feature selection. It can handle non-linear data and give intrinsic feature importance, making it suitable for fingerprinting tasks \cite{breiman2001random}.
    
    \item \textbf{Support Vector Machines (SVM)}: SVM detects an optimal hyperplane that discriminates between classes by maximizing the margin between them; it performs better in binary classification tasks, even with a limited training sample, which makes it ideal for our fingerprinting task of identifying CNN and RNN-based tasks \cite{cortes1995support}.
    
    \item \textbf{Gradient Boosting (XGBoost)}: This is an ensemble learning technique that makes a robust classifier by iteratively combining weaker predictive models \cite{chen2016xgboost}.
\end{itemize}

For each classifier, hyperparameter tuning was conducted using grid-search cross-validation on the training dataset. The performance of classifiers was evaluated on an independent test dataset, employing standard metrics such as accuracy, precision, recall, and F1-score to measure fingerprinting effectiveness comprehensively.

\subsection{Evaluation Metrics}
To evaluate the fingerprinting performance, we use the following widely accepted metrics:

\begin{itemize}
    \item \textbf{Accuracy}: It measures the fraction of correctly predicted instances among all tested instances.
    \item \textbf{Precision}: It quantifies the proportion of true positive predictions among all predicted positives, thus reflecting false-positive sensitivity
    \item \textbf{Recall}: It measures how effectively a classifier identifies actual positive instances, capturing sensitivity to false negatives.
    \item \textbf{F1-score}: It combines precision and recall into a single metric, providing a balanced measure of classification effectiveness, particularly useful when classes are imbalanced or when precision and recall are both crucial.
\end{itemize}

These metrics are calculated and reported separately for each classifier (Random Forest, SVM, and XGBoost) and each class (CNN and RNN), providing comprehensive insight into their respective fingerprinting capabilities.

\subsection{Result Analysis}
%\vspace{-0.3cm}
\begin{table}[htbp]
\centering
\caption{Fingerprint performance for CNN and RNN architectures.}
\label{tab:classification_results}
\vspace{-0.2cm}
\resizebox{\columnwidth}{!}{ % <-- Adjusted width here
\begin{tabular}{|c|c|c|c|c|c|}
\hline
\textbf{Method} & \textbf{Class} & \textbf{Precision} & \textbf{Recall} & \textbf{F1-score} & \textbf{Accuracy} \\ \hline
\multirow{2}{*}{Random Forest} & CNN & 1.00 & 1.00 & 1.00 & \multirow{2}{*}{100\%} \\ \cline{2-5}
                               & RNN & 1.00 & 1.00 & 1.00 & \\ \hline
\multirow{2}{*}{SVM} & CNN & 1.00 & 0.92 & 0.96 & \multirow{2}{*}{95.65\%} \\ \cline{2-5}
                               & RNN & 0.92 & 1.00 & 0.96 & \\ \hline
\multirow{2}{*}{XGBoost} & CNN & 0.92 & 1.00 & 0.96 & \multirow{2}{*}{95.65\%} \\ \cline{2-5}
                               & RNN & 1.00 & 0.91 & 0.95 & \\ \hline
\end{tabular}}
%\vspace{-0.3cm}
\end{table}

Table~\ref{tab:classification_results} summarizes the fingerprinting performance using the well-established three machine learning models. Using these models, we distinguish between the CNN and RNN architectures that we utilize in federated learning. We modeled the fingerprinting methods as binary classification tasks, where each classifier was trained and evaluated using traffic data generated by training CNN and RNN models independently on the client devices in a simulated FL environment. Among the classifiers, Random Forest algorithms show performance by accurately classifying all test instances perfectly, achieving perfect scores across all other metrics.

The SVM classifier gained an overall average accuracy of 95.65\%, misclassifying only one CNN instance as RNN. Regardless of a slight reduction in CNN recall to 92\%, the classifier achieved perfect CNN precision at 100\% and a high overall F1-score of 95.65\%. The XGBoost classifier attained an accuracy of 95.65\%, misclassifying a single RNN instance as CNN. This led to a CNN precision drop to 92\% while maintaining perfect recall at 100\% and a strong F1 score of 95.65\%.

These results confirm the capability of accurately recognizing DL architectures using only network-layer traffic. The performance of Random Forest represents its robustness and suitability for this fingerprinting task, while slight errors observed in SVM and XGBoost suggest a potential sensitivity to noise or overlapping features.

\section{Discussion and Future Directions}

\subsection{Discussion}  
Our study demonstrates that deep learning architectures can be effectively fingerprinted by analyzing network traffic patterns in federated learning environments. By leveraging network-layer metadata and statistical traffic features, our approach successfully distinguishes between CNN and RNN architectures with high accuracy across different classifiers. This finding highlights a significant privacy vulnerability in federated learning, where an adversary can infer critical details about a deployed deep learning model through passive network observation.  

While our results confirm the feasibility of such an attack, they also reveal several limitations. First, our experiments were conducted in a controlled environment where network conditions were ideal. Although we introduced some noise by browsing the web during data collection, real-world federated learning systems operate in far more complex and unpredictable conditions. Network packet loss, retransmissions, congestion, and multiplexed traffic from multiple sources may reduce the effectiveness of our attack.  

Additionally, our fingerprinting approach assumes that only one type of deep learning model is being trained per traffic instance, whereas real-world federated learning environments often involve multiple architectures running simultaneously, making traffic patterns significantly more complex. Another limitation is the dataset size and model diversity. We tested our fingerprinting method on a small sample of deep learning architectures, whereas federated learning scenarios include a much wider variety of deep learning structures. This limited dataset likely led to overfitting to specific traffic patterns, reducing the generalizability of our fingerprinting method.  

Despite these limitations, our results strongly indicate that deep learning architectures exhibit unique network traffic patterns, and with a more robust framework, it should be possible to identify different architectures even in realistic, complex network conditions.  

\subsection{Future Research Directions}  
To enhance the practicality and robustness of our fingerprinting approach, several key research directions should be pursued:  

\begin{itemize}
    \item \textbf{Fingerprinting in Real-World Network Conditions:} Future research should examine how packet loss, congestion, encryption, and background traffic affect fingerprinting accuracy. To manage multiplexed traffic, where many deep learning models train on the same network, models should be refined.

    \item \textbf{Expanding Scope of Attack and Defenses:} Modern federated learning deployments incorporate transformers (BERT, ViTs), RLs, and GANs. Checking if these designs have unique network traffic patterns is crucial. This new risk in federated learning must be mitigated by creating effective countermeasures such as network-layer traffic concealment, adversarial perturbations, and safe aggregation.

    \item \textbf{Building a Scalable and Generalized Framework:} Future research should train fingerprint models on vast, diversified datasets with different architectures, network conditions, and traffic patterns to improve generalizability. Fine-tuned classification models that adapt to real-world federated learning settings and ensure reliable fingerprinting across deployment scenarios will result.
\end{itemize}  

By addressing these areas, future work can enhance attack effectiveness while strengthening security defenses, ensuring greater privacy and robustness in federated learning systems.

\section{Conclusions}
We proposed a novel method to fingerprint deep learning architectures by analyzing network traffic patterns. Our approach consists of a multi-step procedure, including traffic preprocessing, feature engineering, and classification using machine learning techniques. While this study presents an initial exploration, our findings highlight a previously overlooked privacy threat in federated learning environments.

This vulnerability is particularly concerning for safety-critical applications where deep learning models operate as core components, such as autonomous driving and healthcare systems. Our experimental results demonstrate that DL models are susceptible to fingerprinting attacks through encrypted network leakage information, emphasizing the need for both stronger defenses and more robust attack strategies to assess vulnerabilities further. Future research should focus on increasing attack resilience in more diverse network conditions and developing countermeasures to mitigate this security risk effectively.

%%
%% The next two lines define the bibliography style to be used, and
%% the bibliography file.
\bibliographystyle{ACM-Reference-Format}
\bibliography{references}

\end{document}